\documentclass[10pt,twocolumn,letterpaper]{article}

\usepackage{cvpr}              % To produce the CAMERA-READY version
\usepackage{graphicx}
\usepackage{pdflscape}  % or \usepackage{lscape}

\definecolor{cvprblue}{rgb}{0.21,0.49,0.74}
\usepackage[pagebackref,breaklinks,colorlinks,allcolors=cvprblue]{hyperref}

\def\paperID{*****} % *** Enter the Paper ID here
\def\confName{CVPR}
\def\confYear{2026}

\title{PCA-VAE: Differentiable Subspace Quantization without Codebook Collapse
}

\author{
Hao Lu$^{1}$\thanks{Corresponding author: Hao.Lu@advocatehealth.org} \quad
Onur C. Koyun$^{1}$ \quad
Yongxin Guo$^{1}$ \quad
Zhengjie Zhu$^{1}$ \quad
Abbas Alili$^{1}$ \quad
Metin Nafi Gurcan$^{1}$\\[4pt]
$^{1}$Wake Forest University School of Medicine, Winston-Salem, NC, USA\\[4pt]
{\tt\small \{Hao.Lu, Onur.Koyun, Abbas.Alili\}@advocatehealth.org} \\
{\tt\small \{Yongxin.Guo, Zhengjie.Zhu, Metin.Gurcan\}@wfusm.edu}
}

\begin{document}
\maketitle
\begin{abstract}
Vector-quantized autoencoders deliver high-fidelity latents but suffer inherent flaws: the quantizer is non-differentiable, requires straight-through hacks, and is prone to collapse. We address these issues at the root by replacing VQ with a simple, principled, and fully-differentiable alternative: an online PCA bottleneck trained via Oja’s rule. The resulting model, PCA-VAE, learns an orthogonal, variance ordered latent basis without codebooks, commitment losses, or lookup noise. Despite its simplicity, PCA-VAE exceeds VQ-GAN and SimVQ in reconstruction quality on CelebAHQ—while using 10–100× fewer latent bits. It also produces naturally interpretable dimensions (e.g., pose, lighting, gender cues) without adversarial regularization or disentanglement objectives. These results suggest that PCA is a viable replacement for VQ: mathematically grounded, stable, bit-efficient, and semantically structured—offering a new direction for generative models beyond vector quantization.  
The code is available at: \url{https://github.com/CAIR-LAB-WFUSM/OPCA-VAE.git}

\end{abstract}    
\section{Introduction}
\label{sec:intro}

Vector Quantization (VQ) has been a long-standing tool in signal processing for discretizing continuous representations \cite{RN1}. Its integration into deep generative models—most notably through VQ-VAE \cite{RN2}—has proven remarkably successful, demonstrating that inserting a clustering mechanism in the latent space of an autoencoder can substantially improve generative modeling performance. Since then, VQ has become a cornerstone of modern architectures such as VQ-GAN \cite{RN3} and Latent Diffusion Models (LDMs) \cite{RN4}, as well as a fundamental component of vision–language tokenizers \cite{RN5} that bridge visual and linguistic modalities.

Despite its empirical success, VQ is inherently non-differentiable and exhibits several well-known limitations.  
First, the quantization operation in VQ involves a discrete $\arg\min$ selection over the codebook, which blocks gradient flow. Consequently, VQ-VAE must rely on surrogate techniques such as the straight-through estimator (STE) \cite{RN2}, Gumbel-Softmax relaxation \cite{RN6}, or codebook rotation tricks to approximate gradients.  
Second, the codebook update rule in standard VQ only modifies the ``winner'' vector at each step. Non-winning entries remain static, often leading to the \emph{codebook collapse} phenomenon, where large portions of the codebook are never used during training. These two issues—non-differentiability and sparse updates—limit the theoretical soundness and practical stability of VQ-based generative models.

In contrast, Principal Component Analysis (PCA) naturally avoids both problems.  
PCA performs a linear projection onto an orthonormal subspace, which is fully differentiable and globally updated. Unlike VQ, which adjusts one code vector per sample, PCA updates all basis vectors jointly via continuous gradients.  
Therefore, PCA is immune to collapse and provides a smooth, mathematically grounded alternative to vector quantization.  
Moreover, PCA inherently \emph{decouples} the latent factors of variation: its orthogonal basis directions capture statistically independent variance components, aligning with one of the key theoretical motivations of Variational Autoencoders (VAEs)—to disentangle latent semantics.

\begin{figure*}[t]
    \centering
    \includegraphics[width=0.7\linewidth]{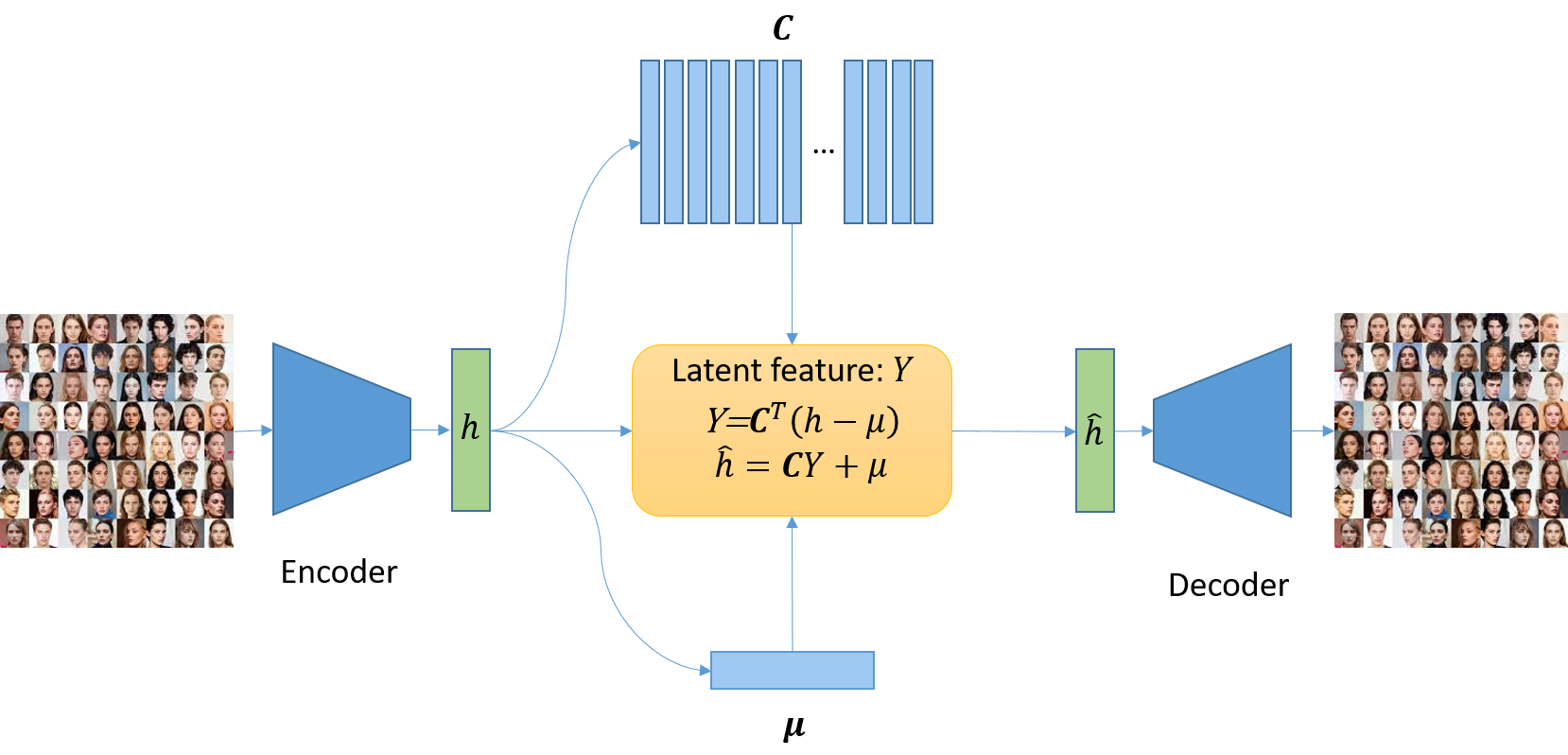}
    \caption{
        \textbf{Overall architecture of the proposed PCA-VAE.}
        The encoder extracts latent features $\mathbf{h}$ from the input image $\mathbf{x}$.
        The PCA layer performs an online orthogonal projection 
        $\hat{\mathbf{h}} = C C^{\top} (\mathbf{h} - \boldsymbol{\mu}) + \boldsymbol{\mu}$,
        where $C$ and $\boldsymbol{\mu}$ are updated via Oja's rule and $r$-fade averaging but treated as stop-gradient variables during VAE backpropagation.
        The quantized latent $\hat{\mathbf{h}}$ is then decoded to reconstruct $\hat{\mathbf{x}}$.
        The PCA layer supports both global (single-vector) and spatial (multi-patch) latent configurations, each with its own PCA basis.
    }
    \label{fig:pca_vae_scheme}
\end{figure*}

Building on these insights, we propose the PCA-VAE, a new generative model that replaces the non-differentiable VQ layer in VQ-VAE with an online PCA module. Specifically, we treat PCA as a trainable layer governed by Oja’s rule, enabling incremental, stable updates to both the mean and basis vectors during training. The resulting framework maintains the expressive power of deep VAEs while achieving orthogonal, continuous, and interpretable latent representations.  
As shown in Fig.~\ref{fig:pca_vae_scheme}, the PCA layer serves as a differentiable projection operator that replaces the discrete quantization stage, thus bridging the gap between discrete clustering and linear subspace learning.

Our contributions are summarized as follows:
\begin{itemize}
    \item We introduce PCA-VAE, a new latent generative model that replaces VQ codebooks with an online-learned PCA layer (trained via Oja’s rule) inside a VAE architecture, enabling end-to-end differentiability and eliminating the need for discrete token learning.

\item We show that PCA-VAE naturally orders and decorrelates latent dimensions by explained variance, yielding axis-aligned semantic factors and highly interpretable latent traversal while achieving competitive or superior reconstruction quality and 10--100$\times$ higher bit-efficiency than VQ-based counterparts.

\end{itemize}

\section{Related Work}

\subsection{VQ-VAE Applications}

Vector Quantization (VQ) is a long-established technique in classical signal processing \cite{gersho1991vector,gray1984vector}.
It represents data by partitioning the input space into discrete regions, each represented by a prototype vector (codeword).  
Mathematically, the VQ process can be viewed as a special case of the $k$-means algorithm \cite{lloyd1982least,macqueen1967some},
where the quantizer’s codebook corresponds to cluster centroids and the encoding step assigns each data vector to its nearest centroid.
Building upon these classical foundations, 
Oord et al. introduced the \textit{Vector-Quantized Variational Autoencoder (VQ-VAE)} \cite{van2017neural},
which integrates a learnable VQ module into the latent bottleneck of an autoencoder.  
This design enables the model to learn discrete latent codes while maintaining a differentiable training pipeline through the straight-through estimator (STE).
The quantized latent representations can then be modelled by powerful autoregressive priors such as PixelCNN or transformers,
resulting in high-quality image, speech, and video synthesis.

The VQ-VAE framework has since become the foundation for many state-of-the-art generative models.  
Razavi et al. proposed VQ-VAE-2 \cite{razavi2019generating}, 
a hierarchical extension that improved the fidelity and diversity of image generation by stacking multiple quantization layers.
VQ tokenization has also been adopted in large-scale text-to-image and diffusion models such as DALL·E and LDMs \cite{ramesh2021zero,RN4},
as well as in video generation frameworks like VideoGPT \cite{yan2021videogpt}.
Moreover, VQ-based discrete representations have become crucial for vision–language models \cite{gu2021vector,esser2021taming},
where images are discretized into sequences of visual tokens that large language models can process.

Overall, these applications demonstrate that incorporating a clustering-based discretization module into the latent space of autoencoders yields compact, compositional, and highly expressive latent representations. VQ-VAE thus bridges the gap between continuous representation learning and discrete token modeling, serving as a fundamental building block for modern generative AI.

\subsection{VQ-VAE Variants}

Some variants have been proposed to address the fundamental
limitations of standard VQ, particularly its non-differentiable
$\arg\min$ quantization and the ``winner-takes-all'' codebook update
that often leads to codebook collapse.  
Unlike classical VQ, which can be viewed as a stochastic approximation
to $k$-means clustering, most of these modern variants no longer adhere
to that geometric interpretation.  
Instead, they redesign the quantization mechanism to improve stability,
gradient flow, or representational flexibility.

For instance, Finite Scalar Quantization (FSQ)~\cite{mentzer2023finite}
dispenses with the notion of a vector codebook entirely.  
It directly quantizes each latent feature dimension into a small number
of scalar levels, effectively replacing the nearest-neighbor lookup with
element-wise discretization.  
This design removes the need for $\arg\min$ operations and eliminates
codebook competition, yielding a fully differentiable quantization
process.

Similarly, Latent Feature Quantization (LFQ) \cite{RN7}
introduces a directional or sign-based quantization scheme, where
latent vectors are normalized to unit length and then discretized by
their orientations rather than Euclidean distances.
This can be interpreted as a form of directional clustering rather than
traditional $k$-means, enabling better angular consistency across latent
codes.

Other approaches modify the structure or dynamics of the codebook itself.
For example, Lattice VQ and Product VQ methods
\cite{khalil2024learnable,vali2023stochastic}
introduce parameterized or factorized codebooks that ensure more uniform
utilization, while Residual Quantization~\cite{RN12}
and Hierarchical VQ-VAE-2 \cite{razavi2019generating}
extend the quantization across multiple levels of abstraction to capture
fine-grained features.

Collectively, modern quantization
modules preserve the core intuition of compressing latent spaces into
finite sets but abandon the strict k-means assumptions that initially 
motivated vector quantization.

\subsection{Online and Streaming PCA}  
In a parallel vein, the literature on principal component analysis (PCA) in streaming or online settings provides a mature theoretical and practical framework for continuous subspace learning. The work on streaming PCA by Mitliagkas et al. \cite{mitliagkas2014memory} offers memory-limited streaming algorithms with convergence guarantees, while broader surveys on subspace tracking appear in Balzano et al. \cite{balzano2018streaming}.  
The original Erkki Oja’s rule~\cite{oja1982simplified} for online PCA has been extended, for example in the convergence analysis of Oja’s algorithm by Xin Liang \cite{liang2021optimality}.  
These methods update basis vectors continuously (rather than using  winner‐takes‐all) and are fully differentiable linear operations.  
Crucially, they avoid the non‐differentiability and codebook collapse issues seen in vector quantization. Despite this, the use of PCA as a latent-layer within deep generative autoencoders remains under‐explored.  
Our work bridges this gap by integrating an online PCA layer—learned via Oja’s algorithm—directly into a generative VAE framework.

\section{Methods}
\subsection{Preliminary Knowledge: Principal Component Analysis}

Principal Component Analysis (PCA) is a classical linear method 
for finding a low-dimensional orthonormal subspace that captures 
the maximal variance of a high-dimensional dataset.
Let 
\(\mathbf{z}_i \in \mathbb{R}^N\) 
denote the $i$-th sample from a centered distribution 
with covariance matrix 
\(\Sigma = \mathbb{E}[\mathbf{z}\mathbf{z}^\top]\).
The goal of PCA is to find 
\(Q\) orthonormal directions 
\(\{\mathbf{c}_1, \ldots, \mathbf{c}_Q\}\)
that maximize the total projected variance:
\begin{align}
\max_{C \in \mathbb{R}^{N\times Q}}
\quad 
\mathrm{tr}\!\left(C^\top \Sigma C\right)
\quad
\text{s.t. } 
C^\top C = I_Q ,
\label{eq:pca_trace_obj}
\end{align}
where \(C = [\mathbf{c}_1, \ldots, \mathbf{c}_Q]\).
Each column \(\mathbf{c}_q\) is referred to as a 
\emph{principal component}.
The solution of \eqref{eq:pca_trace_obj}
is given by the eigenvectors of \(\Sigma\) 
corresponding to its $Q$ largest eigenvalues 
\(\lambda_1 \ge \lambda_2 \ge \cdots \ge \lambda_N\).
These eigenvalues quantify the amount of variance explained by each component.

For a centered input sample \(\mathbf{z}\),
the projection and reconstruction under PCA are
\begin{align}
\mathbf{y} &= C^\top \mathbf{z}, \label{eq:pca_projection}\\
\widehat{\mathbf{z}} &= C\mathbf{y} = C C^\top \mathbf{z}. \label{eq:pca_reconstruction}
\end{align}
The reconstruction error in mean square form is
\begin{align}
\mathcal{L}(C)
&= \mathbb{E}\!\left[\|\mathbf{z} - CC^\top \mathbf{z}\|_2^2\right]
= \mathrm{tr}(\Sigma) - \mathrm{tr}\!\left(C^\top \Sigma C\right). \label{eq:pca_loss_pre}
\end{align}
Minimizing \eqref{eq:pca_loss_pre} is equivalent to the maximization in 
\eqref{eq:pca_trace_obj}.
Hence, the optimal subspace 
\(\mathrm{span}(C)\)
preserves the largest possible variance of the data 
while minimizing the average reconstruction error.

In online or streaming settings,
the true covariance \(\Sigma\) is unknown;
instead, it is estimated from minibatches of samples.
Online PCA algorithms, such as Oja’s rule,
iteratively update \(C\) using stochastic gradients
that approximate the stationary solution of 
Eq.~\eqref{eq:pca_trace_obj}.
This motivates the learning formulation adopted
in our PCA quantizer.

\subsection{Online PCA}\label{sec:classic_pca}
\subsubsection{Notation and setup}

Let a minibatch of latent feature maps be 
\(x \in \mathbb{R}^{B\times D\times H\times W}\)
with batch size \(B\).
We flatten each sample to 
\(\mathrm{vec}(\mathbf{x}_b) \in \mathbb{R}^{DHW}\).
When enabled, a lightweight mapper 
\(\phi:\mathbb{R}^{DHW}\!\to\!\mathbb{R}^{N}\) 
and a decoder 
\(\psi:\mathbb{R}^{N}\!\to\!\mathbb{R}^{DHW}\)
map to and from an intermediate space of dimension \(N\).
The subspace basis is represented by
\begin{align}
C = [\mathbf{c}_1, \ldots, \mathbf{c}_Q] \in \mathbb{R}^{N\times Q},
\quad C^\top C = I_Q,
\end{align}
where each column \(\mathbf{c}_q\) is an orthonormal component.

For each sample \(b\),
\begin{align}
\mathbf{z}_b &=\mathrm{vec}(\mathbf{x}_b)\\
\tilde{\mathbf{z}}_b &= \mathbf{z}_b - \boldsymbol{\mu}.
\end{align}

Projection and reconstruction in the PCA space are given by
\begin{align}
\mathbf{y}_b &= C^\top \tilde{\mathbf{z}}_b, \label{eq:y_proj}\\
\widehat{\mathbf{z}}_b &= C\mathbf{y}_b + \boldsymbol{\mu}. \label{eq:z_rec}
\end{align}
If the mapper is enabled, the reconstructed image is
\(\widehat{\mathbf{x}}_b = \mathrm{reshape}\!\big(\psi(\widehat{\mathbf{z}}_b), D, H, W\big)\);
otherwise, we reshape \(\widehat{\mathbf{z}}_b\) directly.

\subsubsection{Running-mean centering with geometric \texorpdfstring{$\gamma$}{}-fade}

To stabilize the subspace updates, we maintain a running mean 
\(\boldsymbol{\mu}_n \in \mathbb{R}^N\)
of the latent vectors in the PCA space.
Instead of the classical exponential moving average (EMA),
we employ a \emph{geometric fading mean}, or \emph{$\gamma$-fade average},
which assigns exponentially decaying weights \(\gamma^k\) 
to past batch means.

Let the batch mean at step \(t\) be
\begin{align}
\bar{\mathbf{z}}_t = \frac{1}{B}\sum_{b=1}^{B} \mathbf{z}_b^{(t)} .
\end{align}
The cumulative $\gamma$-fade mean after \(n{+}1\) updates is defined as
\begin{align}
\boldsymbol{\mu}_{n+1} 
= 
\frac{ \displaystyle \sum_{k=0}^{n} \gamma^k\,\bar{\mathbf{z}}_{n+1-k} }
{ \displaystyle \sum_{k=0}^{n} \gamma^k },
\qquad \gamma \in (0,1). \label{eq:r_fade_def}
\end{align}
This normalized geometric weighting ensures that 
recent batches dominate when \(\gamma<1\),
while older statistics decay exponentially.

For practical streaming updates,
Eq.~\eqref{eq:r_fade_def} can be implemented recursively as
\begin{align}
\boldsymbol{\mu}_{n+1}
&= \frac{1}{\rho_{n+1}}\bar{\mathbf{z}}_{n+1}
+ \frac{\gamma \rho_n}{\rho_{n+1}}\boldsymbol{\mu}_n, \label{eq:r_fade_recursion}\\
\rho_{n}
&= \frac{1 - \gamma^{n}}{1 - \gamma}
\label{eq:r_fade_coeff}
\end{align}
Thus, the update smoothly transitions between the newest batch mean 
and the accumulated average, 
while guaranteeing normalization by the cumulative geometric weight.
We refer to this process as the \textbf{$\gamma$-fade update}, 
which replaces the bias-corrected EMA used in standard online PCA formulations.

\subsubsection{Online subspace learning (Oja-type update)}

Given a centered minibatch matrix 
\(\tilde{Z} \in \mathbb{R}^{B\times N}\)
whose rows are \(\tilde{\mathbf{z}}_b^\top\),
we compute the projected representations and their Gram matrix:
\begin{align}
Y &= \tilde{Z} C \in \mathbb{R}^{B\times Q},\\
G &= Y^\top Y \in \mathbb{R}^{Q\times Q}.
\end{align}
Let \(\mathrm{Up}(G)\) denote the elementwise upper-triangular part of \(G\)
(including the diagonal).
The Oja-type update with learning rate \(\eta_t>0\) is
\begin{align}
\Delta C &= \frac{1}{B}\!\left(\tilde{Z}^\top Y - C\,\mathrm{Up}(G)\right), \label{eq:dC}\\
C &\leftarrow C + \eta_t\,\Delta C. \label{eq:C_update}
\end{align}
The first term in Eq.~\eqref{eq:dC} performs stochastic ascent
on the variance captured by the subspace,
while the second term \(-C\,\mathrm{Up}(G)\)
damps within-subspace rotations, improving numerical stability.

\subsubsection{Symmetric re-orthonormalization}

To prevent drift from numerical accumulation, 
we periodically re-orthonormalize \(C\) using the symmetric inverse square root of its Gram matrix:
\begin{align}
G_C &= C^\top C = U\Lambda U^\top,\\
\Lambda_\epsilon &= \mathrm{diag}\!\big(\max(\lambda_i,\epsilon)\big),\\
S^{-1/2} &= U\,\Lambda_\epsilon^{-1/2} U^\top,\\
C &\leftarrow C\,S^{-1/2}. \label{eq:orthonorm}
\end{align}
This ensures \(C^\top C \approx I_Q\)
while minimally perturbing the learned span.

\subsubsection{Objective and optimality (link to PCA)}

For a zero-mean random vector \(\mathbf{z}\in\mathbb{R}^{N}\)
with covariance 
\(\Sigma = \mathbb{E}[\mathbf{z}\mathbf{z}^\top]\),
the Oja-type update in Eqs.~\eqref{eq:dC}--\eqref{eq:C_update},
together with the orthonormalization in Eq.~\eqref{eq:orthonorm},
forms a stochastic online algorithm that converges 
to the optimal solution of Eq.~\ref{eq:pca_loss_pre} \cite{oja1982simplified}.

\subsection{Integration of PCA Layer into VAE}

In our PCA-VAE, the conventional latent bottleneck of a Variational Autoencoder 
is replaced by an online-learned PCA quantization layer. Figure~\ref{fig:pca_vae_scheme} illustrates the overall pipeline of the proposed PCA-VAE, 
where the PCA layer replaces the stochastic sampling of a traditional VAE with an online orthogonal projection step.
This design preserves the low-dimensional structure of the latent representation 
through orthogonal projection while allowing the PCA basis 
to adapt dynamically to the feature distribution during training.

\subsubsection{Architecture overview}

Let the encoder network 
\(E_\theta : \mathbb{R}^{C\times H\times W} \!\to\! \mathbb{R}^{D\times H'\times W'}\)
map an input image \(\mathbf{x}\) into latent features
\(\mathbf{h} = E_\theta(\mathbf{x})\).
These features are then passed into a PCA quantizer 
\(Q_\phi\) (the PCA layer), producing the quantized latent code
\begin{align}
\widehat{\mathbf{h}} = Q_\phi(\mathbf{h}; \text{sg}(C), \text{sg}(\boldsymbol{\mu})),
\end{align}
where \(C\in\mathbb{R}^{N\times Q}\) denotes the learned orthonormal basis
and \(\boldsymbol{\mu}\in\mathbb{R}^{N}\) is the running mean
estimated using the geometric $r$-fade update
defined in Eq.~\eqref{eq:r_fade_def}.
The reconstructed image is obtained by decoding
\(\widehat{\mathbf{h}}\) through the decoder network 
\(D_\psi : \mathbb{R}^{D\times H'\times W'}\!\to\!\mathbb{R}^{C\times H\times W}\):
\begin{align}
\widehat{\mathbf{x}} = D_\psi(\widehat{\mathbf{h}}).
\end{align}
The model is trained to minimize the reconstruction loss:
\begin{align}
\mathcal{L}_{\text{PCA-VAE}}
= \|\widehat{\mathbf{x}} - \mathbf{x}\|_2^2.
\end{align}

\subsubsection{Stop-gradient treatment of PCA parameters}

During VAE training, both \(C\) and \(\boldsymbol{\mu}\) 
are updated using the Oja-type and $r$-fade rules 
described in Sec.~\ref{sec:classic_pca}.
However, these updates are \emph{outside} the standard 
VAE backpropagation graph.
In other words, \(C\) and \(\boldsymbol{\mu}\)
are treated as \textbf{stop-gradient (sg)} variables during the encoder--decoder optimization:
\begin{align}
\frac{\partial \mathcal{L}_{\text{PCA-VAE}}}{\partial C} = 0, 
\qquad
\frac{\partial \mathcal{L}_{\text{PCA-VAE}}}{\partial \boldsymbol{\mu}} = 0.
\end{align}
This design ensures that the subspace learning dynamics of the PCA layer
follow the stable Oja updates,
while the encoder and decoder are trained purely 
through the reconstruction loss.
In effect, the PCA layer acts as a non-parametric, 
self-organizing quantizer embedded in the VAE pipeline.

\subsubsection{Single-vector and multi-patch latent structures}

Depending on the latent layout,
the PCA-VAE can operate in two configurations:

\begin{enumerate}[label=(\roman*)]
\item \textbf{Single-vector latent:}
The encoder output \(\mathbf{h}\in\mathbb{R}^{D\times H'\times W'}\)
is flattened into one global feature vector 
\(\mathbf{z}\in\mathbb{R}^{N}\)
before being projected onto a shared PCA basis \(C\).
This setting captures global semantic factors in the image.

\item \textbf{Multi-patch latent:}
The encoder output is divided into spatial patches
\(\mathbf{h}_{(p)} \in \mathbb{R}^{D}\) 
for \(p=1,\ldots,P\), where \(P = H'W'\).
Each patch has an independent PCA basis \(C_p\in\mathbb{R}^{D\times Q}\)
and running mean \(\boldsymbol{\mu}_p\),
updated separately using the local feature statistics:
\begin{align}
\widehat{\mathbf{h}}_{(p)} = 
C_p C_p^\top (\mathbf{h}_{(p)} - \boldsymbol{\mu}_p)
+ \boldsymbol{\mu}_p.
\end{align}
This configuration allows localized, position-dependent
feature compression analogous to spatial quantization
in VQ-VAE, but with orthogonal linear projections.
\end{enumerate}

\subsubsection{Summary of integration}

In summary, the PCA layer replaces stochastic sampling in a conventional VAE
with an \emph{online orthogonal projection} stage.
The parameters \((C,\boldsymbol{\mu})\) are updated analytically via Oja’s rule 
and geometric $r$-fade averaging,
while gradients from the reconstruction loss 
propagate only through the encoder and decoder.
This hybrid design combines the statistical interpretability of PCA 
with the nonlinear modeling power of deep VAEs,
supporting both global (single-vector) and spatial (multi-patch) latent forms.

\section{Experiments}
In this work, we focus exclusively on the reconstruction task in order to isolate the effect of replacing the VQ bottleneck with our online PCA layer.
All reported metrics (rFID, SSIM, LPIPS) are computed between inputs and their reconstructions produced by the PCA-VAE.

Although our experiments emphasize reconstruction, the PCA-coded latents are \emph{continuous} and can be integrated into modern generative pipelines.
For latent diffusion models (e.g., LDM~\cite{RN4}), the PCA coefficients can serve directly as the latent space or as conditioning features without requiring discretization.

We adopt the CelebA-HQ (256×256) dataset, which contains 30,000 high-quality celebrity face images resampled to 256 pixels. Initially introduced by NVIDIA \cite{RN17}, this dataset is widely used for generative modeling and contains only images without attributes or labels. 
We employ MSE loss as the reconstruction objective, and evaluate reconstruction quality using rFID, LPIPS, and SSIM metrics. We set batch size as 16, learning rate as 0.0005, use Adam optimizer.

\section{Results}
\subsection{Multi-Metric Reconstruction Comparison}

We compare the strongest configuration of our method, PCA-VAE with a 16$\times$16 latent grid and 
100\% orthogonal bases, against VQGAN~\cite{RN3}, SimVQ~\cite{RN13} (SOTA VQVAE in 2024), VQ-VAE~\cite{razavi2019generating}, and an AutoencoderKL~\cite{RN4} (VAE) baseline. 
All VQ models use a 16$\times$16 latent grid with an equal token budget of 8,912 codebook indices. 
AutoencoderKL~\cite{RN4} (VAE) also operates on a 16$\times$16 latent grid with a Gaussian latent prior.

\begin{figure}[t]
    \centering
    \includegraphics[width=0.70\linewidth]{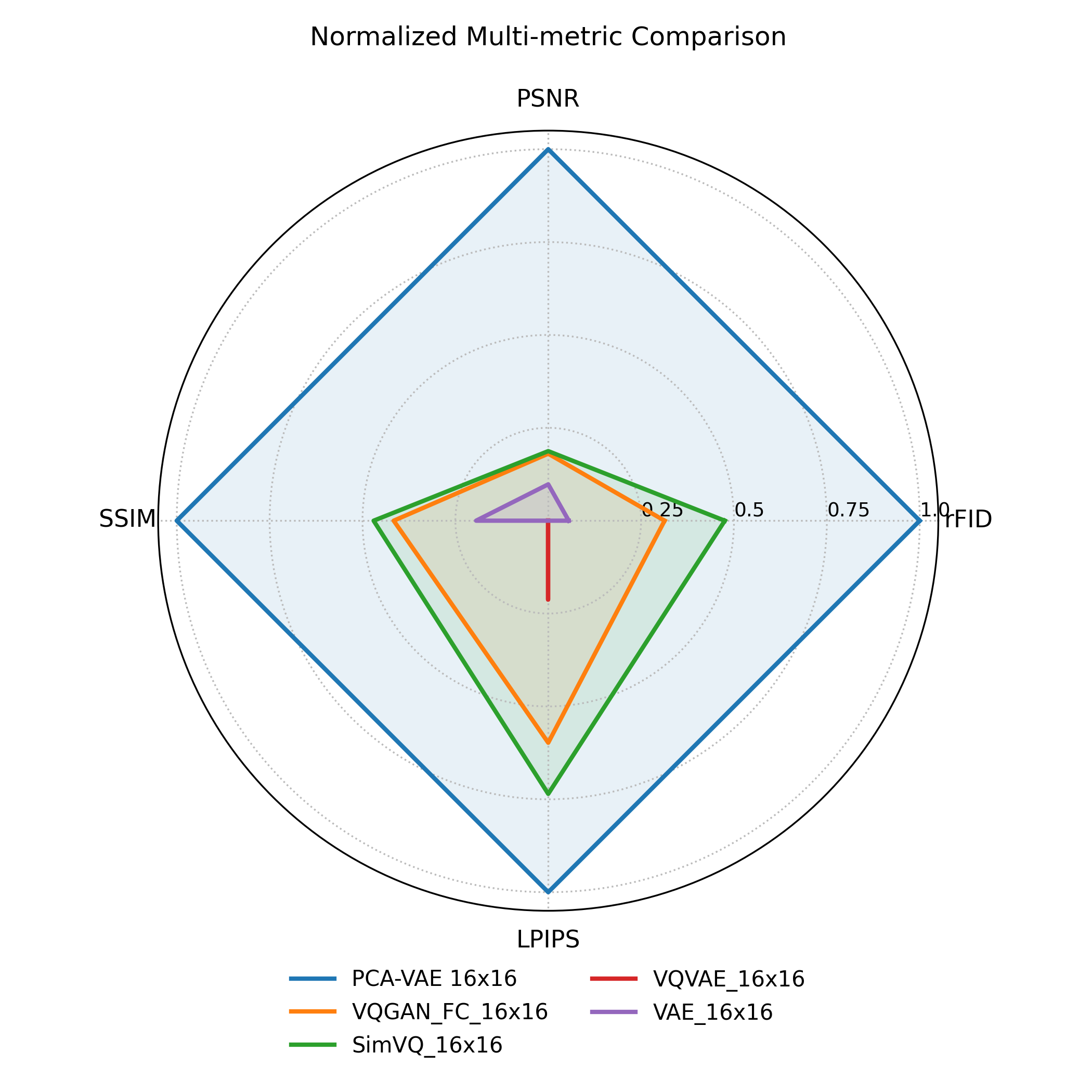}
    \caption{
    Normalized reconstruction performance. 
    We compare PCA-VAE (16$\times$16, 100\% bases) with VQGAN~\cite{RN3}, SimVQ~\cite{RN13}, VQ-VAE~\cite{razavi2019generating}, and a VAE~\cite{RN4} baseline. 
    All VQ models use 16$\times$16 latents and 8,912 codebook tokens. 
    Metrics are normalized (PSNR/SSIM min–max, LPIPS/rFID reverse) so that higher is better. 
    }
    \label{fig:radar}
\end{figure}

We report PSNR, SSIM, LPIPS, and rFID. PSNR and SSIM are min–max normalized, while LPIPS and rFID use 
reverse min–max normalization so that higher scores uniformly indicate better performance. 
Figure~\ref{fig:radar} shows that PCA-VAE attains the most balanced and overall highest reconstruction 
quality across all four metrics, outperforming discrete codebook approaches despite using a continuous latent space.

\noindent\textit{Clarification.}
Discrete VQ models quantize latent features into codebook indices. PCA-VAE, in contrast, uses continuous orthogonal basis coefficients. 
To ensure fairness, all VQ models share the same spatial latent resolution and token budget (16$\times$16, 8,912 tokens), 
and AutoencoderKL serves as a continuous latent baseline under the same latent grid configuration.

\subsection{Orthogonal Basis Efficiency and Scaling Behavior}

\begin{figure*}[t]
\centering
\begin{subfigure}{0.42\linewidth}
    \includegraphics[width=\linewidth]{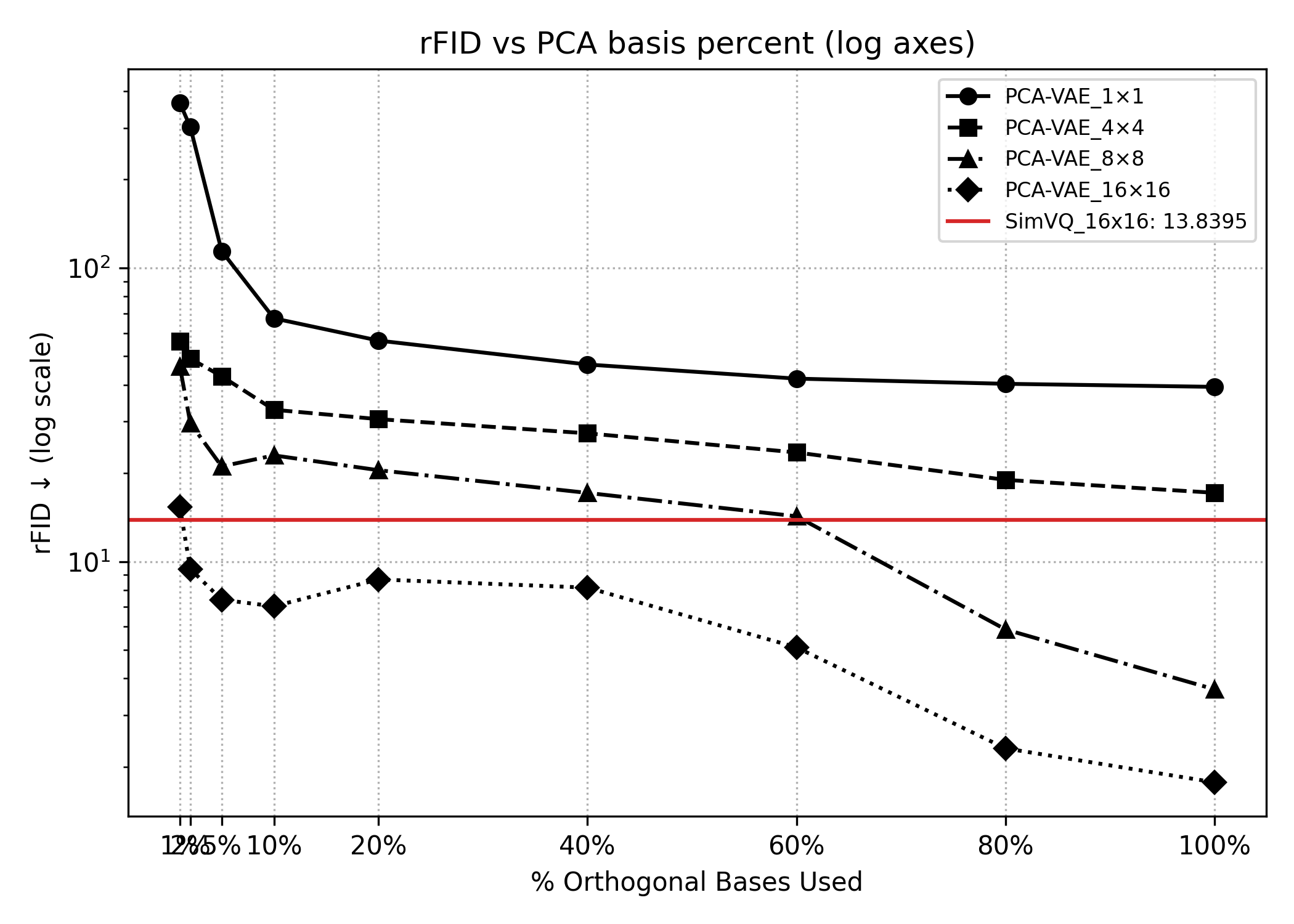}
    \caption{rFID (log scale)}
\end{subfigure}
\hspace{0.02\linewidth}
\begin{subfigure}{0.42\linewidth}
    \includegraphics[width=\linewidth]{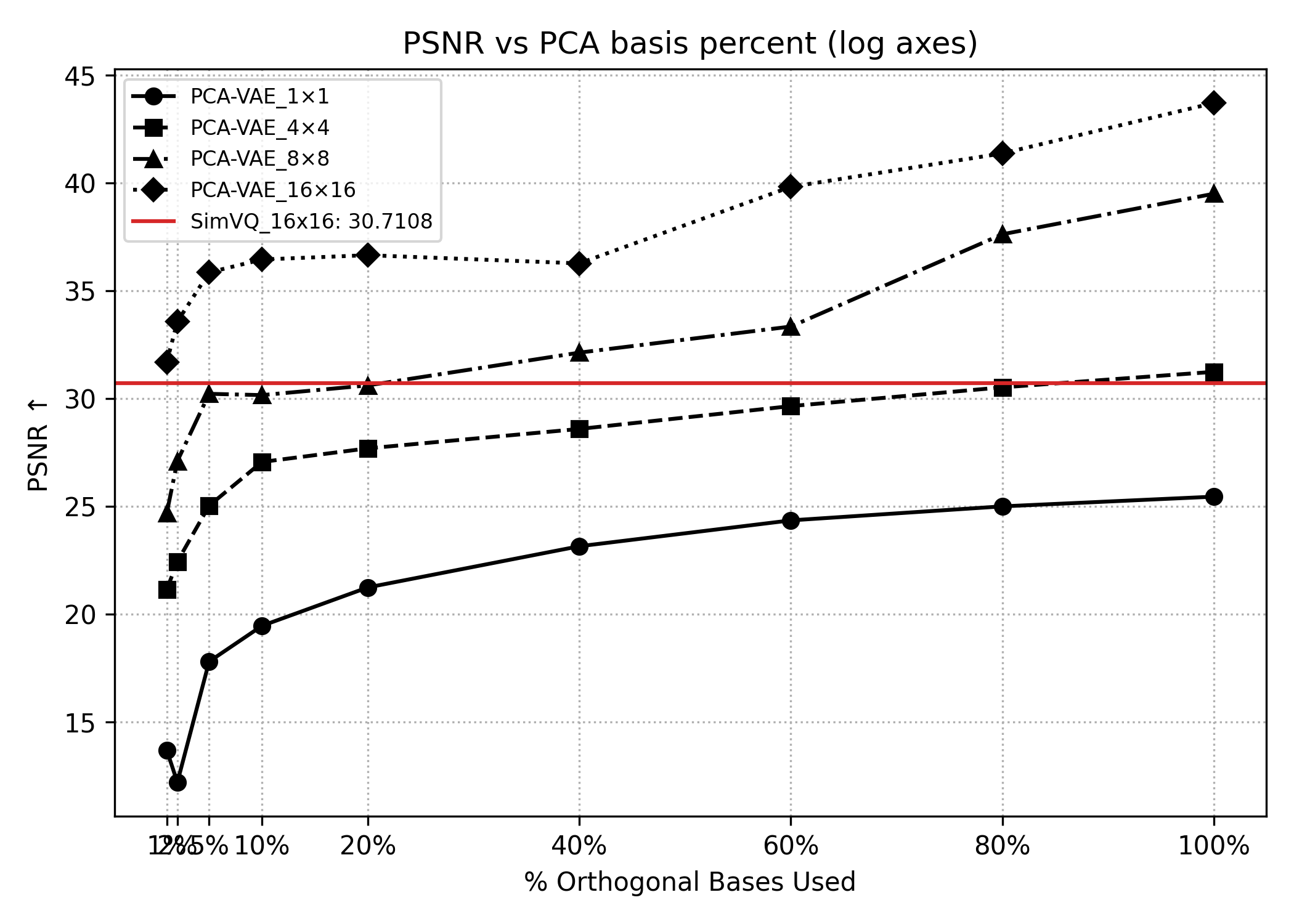}
    \caption{PSNR}
\end{subfigure}
\\[4pt]
\begin{subfigure}{0.42\linewidth}
    \includegraphics[width=\linewidth]{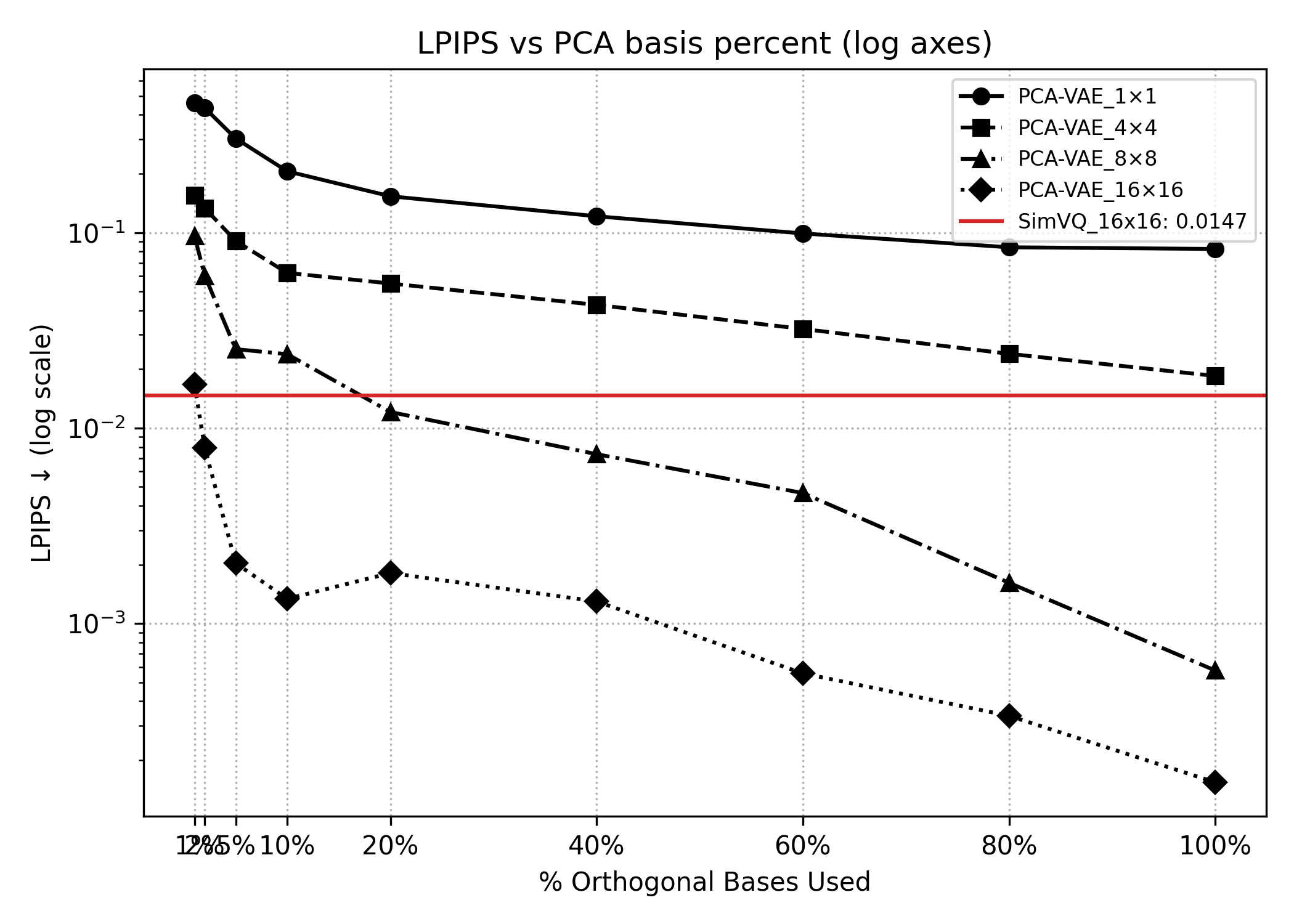}
    \caption{LPIPS (log scale)}
\end{subfigure}
\hspace{0.02\linewidth}
\begin{subfigure}{0.42\linewidth}
    \includegraphics[width=\linewidth]{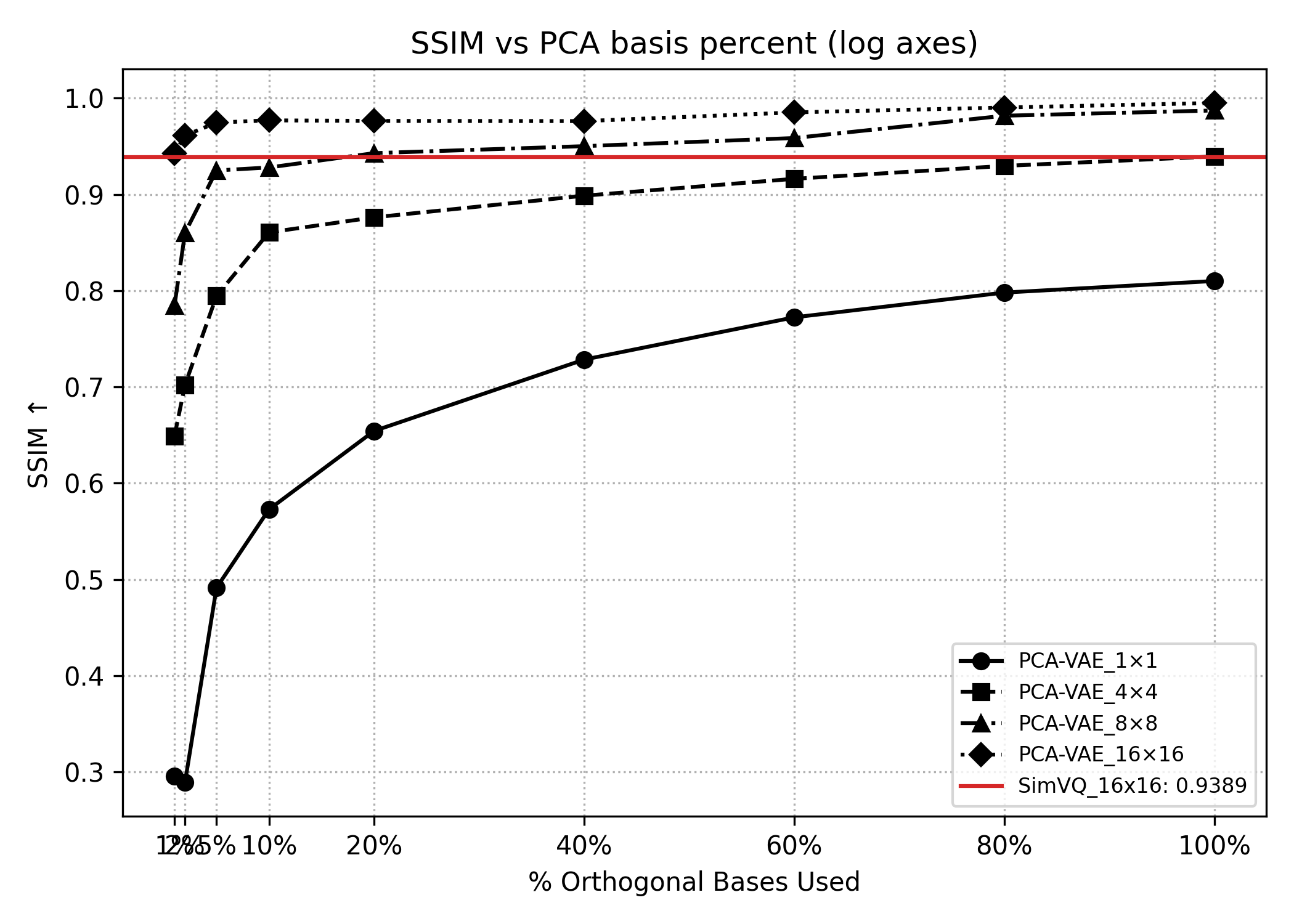}
    \caption{SSIM}
\end{subfigure}
\caption{
Scaling behavior of PCA-VAE with respect to the fraction of principal bases used (1\%--100\%)
under different latent grid resolutions (1$\times$1, 4$\times$4, 8$\times$8, 16$\times$16).
The red horizontal line marks the best VQ baseline (SimVQ~\cite{RN13}, 16$\times$16). 
}
\label{fig:pca_scaling}
\end{figure*}

We investigate how reconstruction quality scales with both latent grid resolution 
(1$\times$1, 4$\times$4, 8$\times$8, 16$\times$16) and the fraction of PCA bases retained 
from the 256-dimensional latent (1\% to 100\%). 
Figure~\ref{fig:pca_scaling} reports SSIM, PSNR, LPIPS, and rFID as we gradually increase 
the number of orthogonal bases. For LPIPS and rFID, log axes are used due to long-tailed behavior. 
The best-performing VQ baseline (SimVQ~\cite{RN13} at a 16$\times$16 grid) is marked as a horizontal reference.

Across all metrics, PCA-VAE exhibits smooth and monotonic scaling: 
larger latent grids and more principal components consistently improve performance. 
Notably, PCA-VAE approaches or exceeds the SimVQ~\cite{RN13} baseline using only 5--10\% of the PCA bases, 
indicating that most perceptual signal energy is concentrated in the top eigen-directions. 
Even with aggressive basis truncation, PCA-VAE outperforms discrete codebook methods in the low-dimensional regime.

\subsection{Latent Bit--Budget Efficiency}

To evaluate representational efficiency, we measure reconstruction quality as a
function of the latent bit budget, defined as the total number of bits required
to store the latent representation of an image. For continuous latent models
such as PCA-VAE and AutoencoderKL, the latent budget equals
$N \times k \times b$, where $N$ is the number of spatial latent tokens,
$k$ is the number of active latent channels (i.e., retained PCA bases),
and $b = 32$ bits per floating point value. For discrete vector--quantized models
such as VQGAN~\cite{RN3}, SimVQ~\cite{RN13}, and VQ-VAE~\cite{razavi2019generating}, the budget equals
$N \times \log_2 K$, where $K$ is the codebook size and each token stores an index.

\begin{figure*}[t]
\centering
\begin{subfigure}{0.42\linewidth}
    \includegraphics[width=\linewidth]{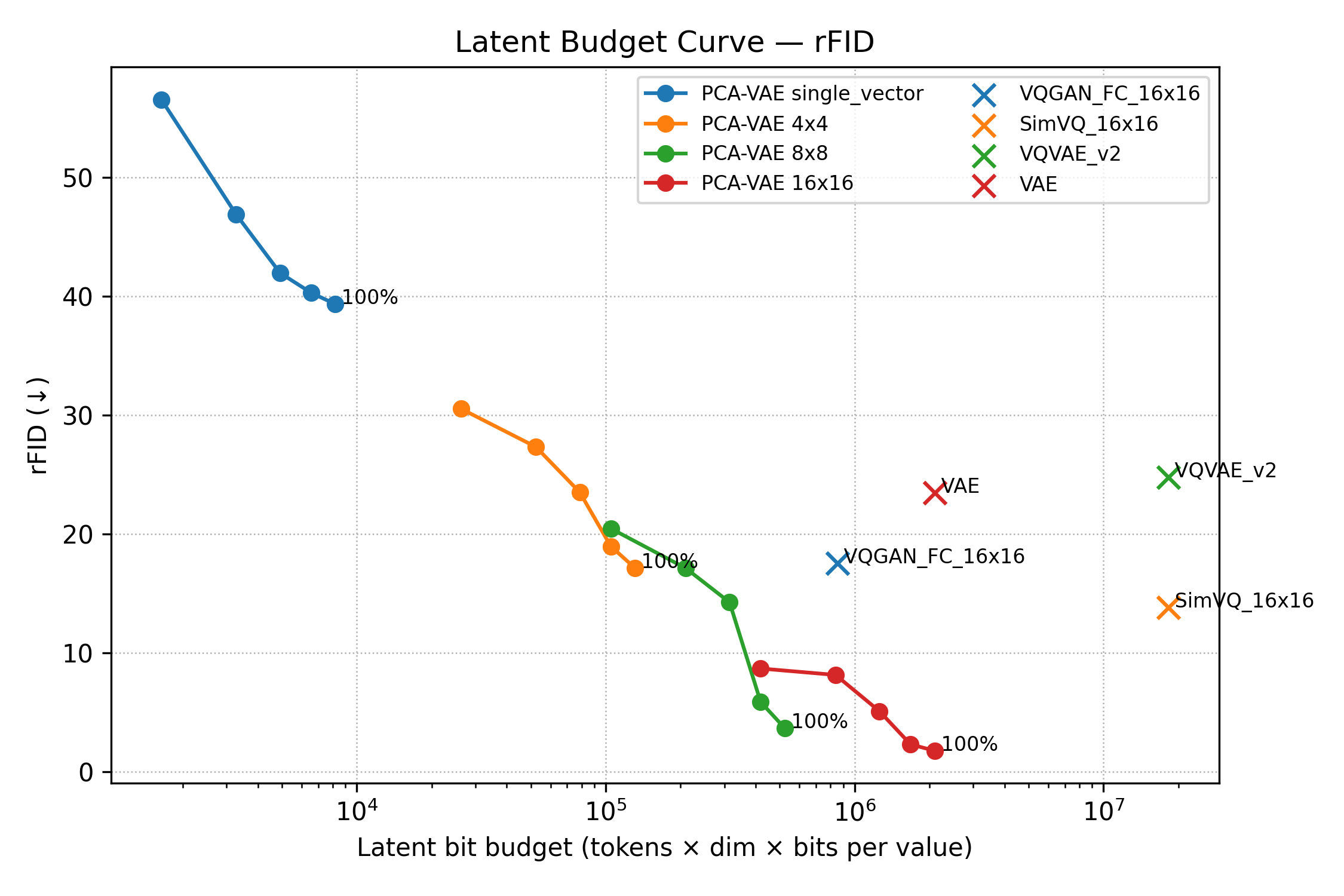}
    \caption{rFID (lower is better)}
\end{subfigure}
\hspace{0.02\linewidth}
\begin{subfigure}{0.42\linewidth}
    \includegraphics[width=\linewidth]{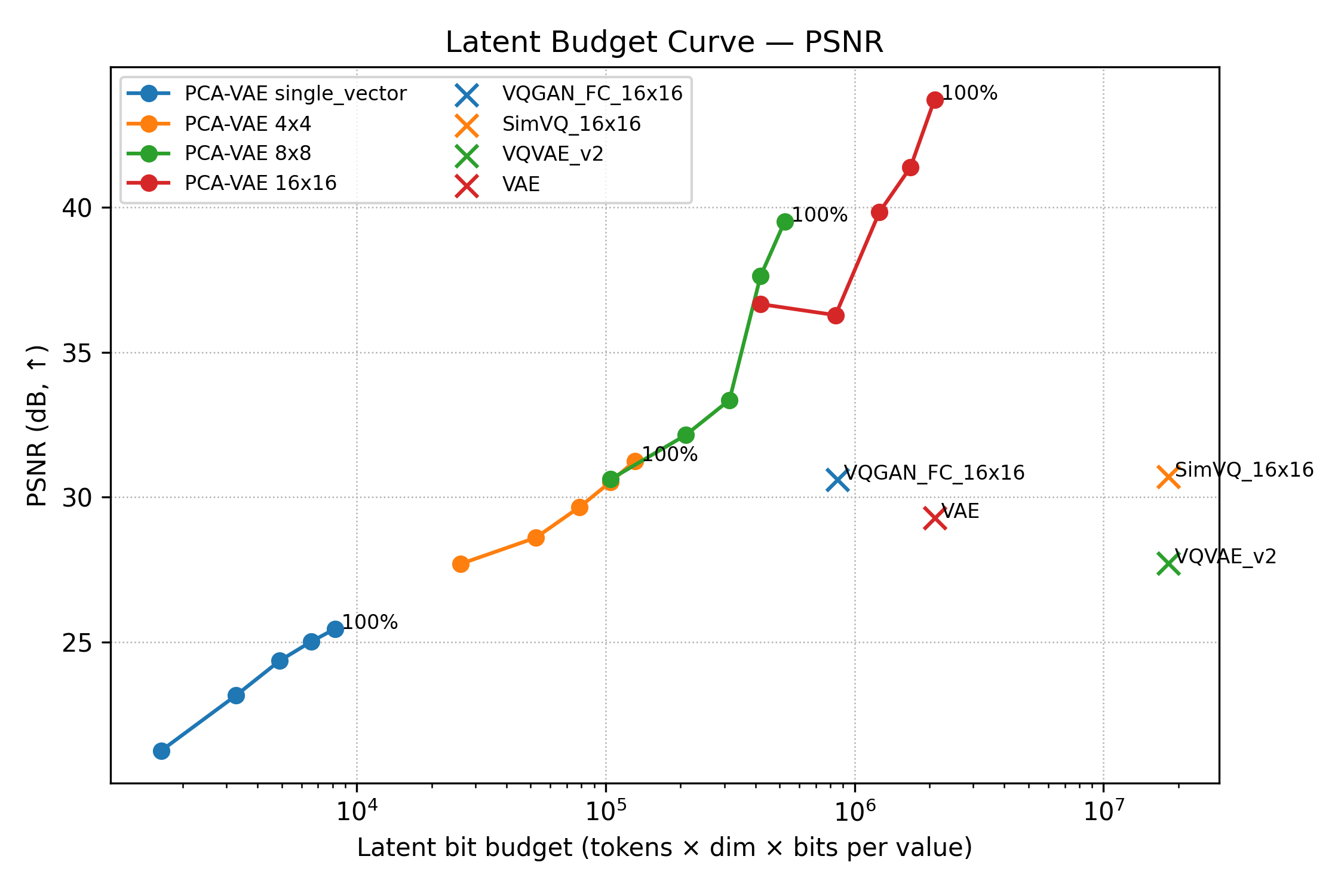}
    \caption{PSNR (higher is better)}
\end{subfigure}
\\[4pt]
\begin{subfigure}{0.42\linewidth}
    \includegraphics[width=\linewidth]{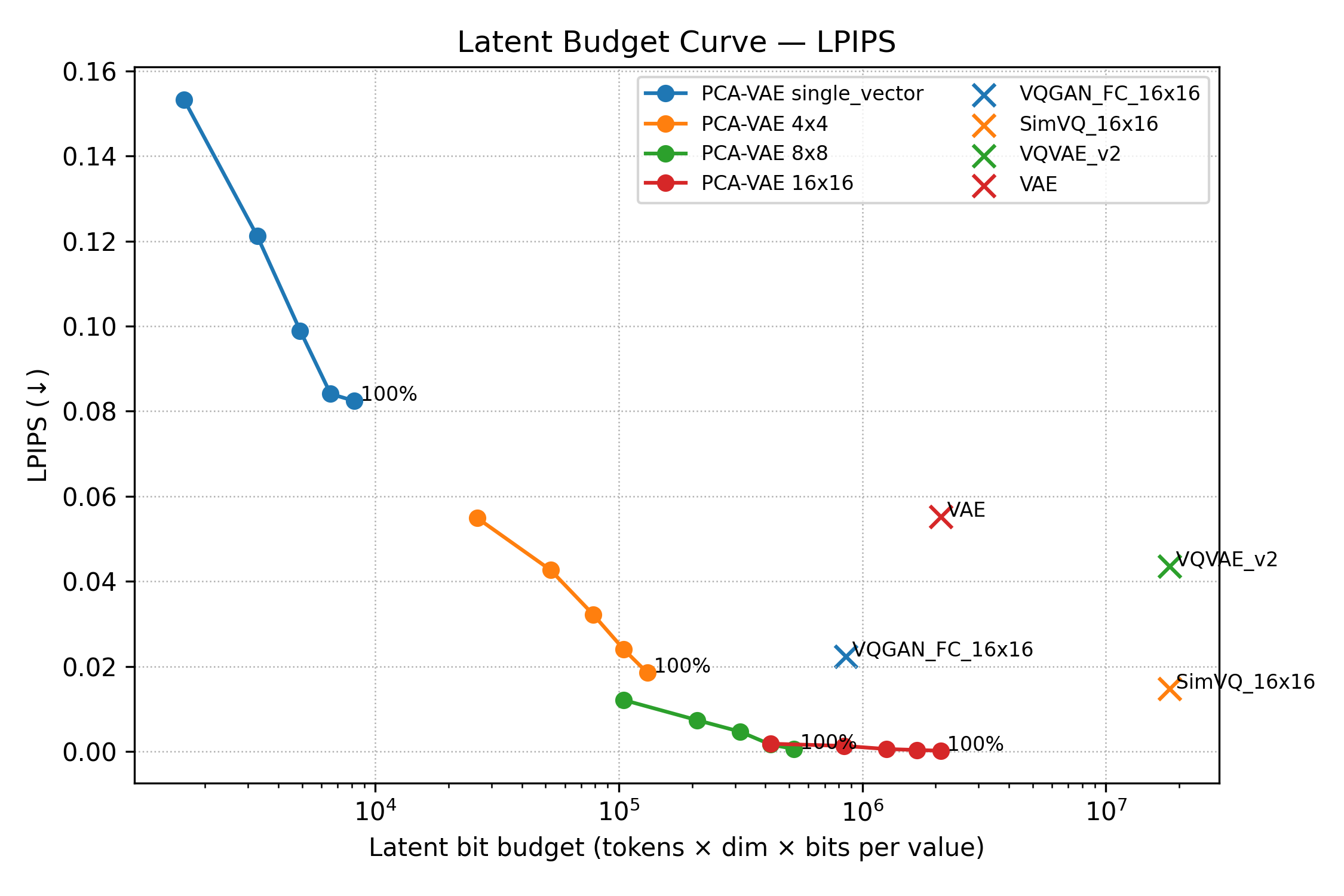}
    \caption{LPIPS (lower is better)}
\end{subfigure}
\hspace{0.02\linewidth}
\begin{subfigure}{0.42\linewidth}
    \includegraphics[width=\linewidth]{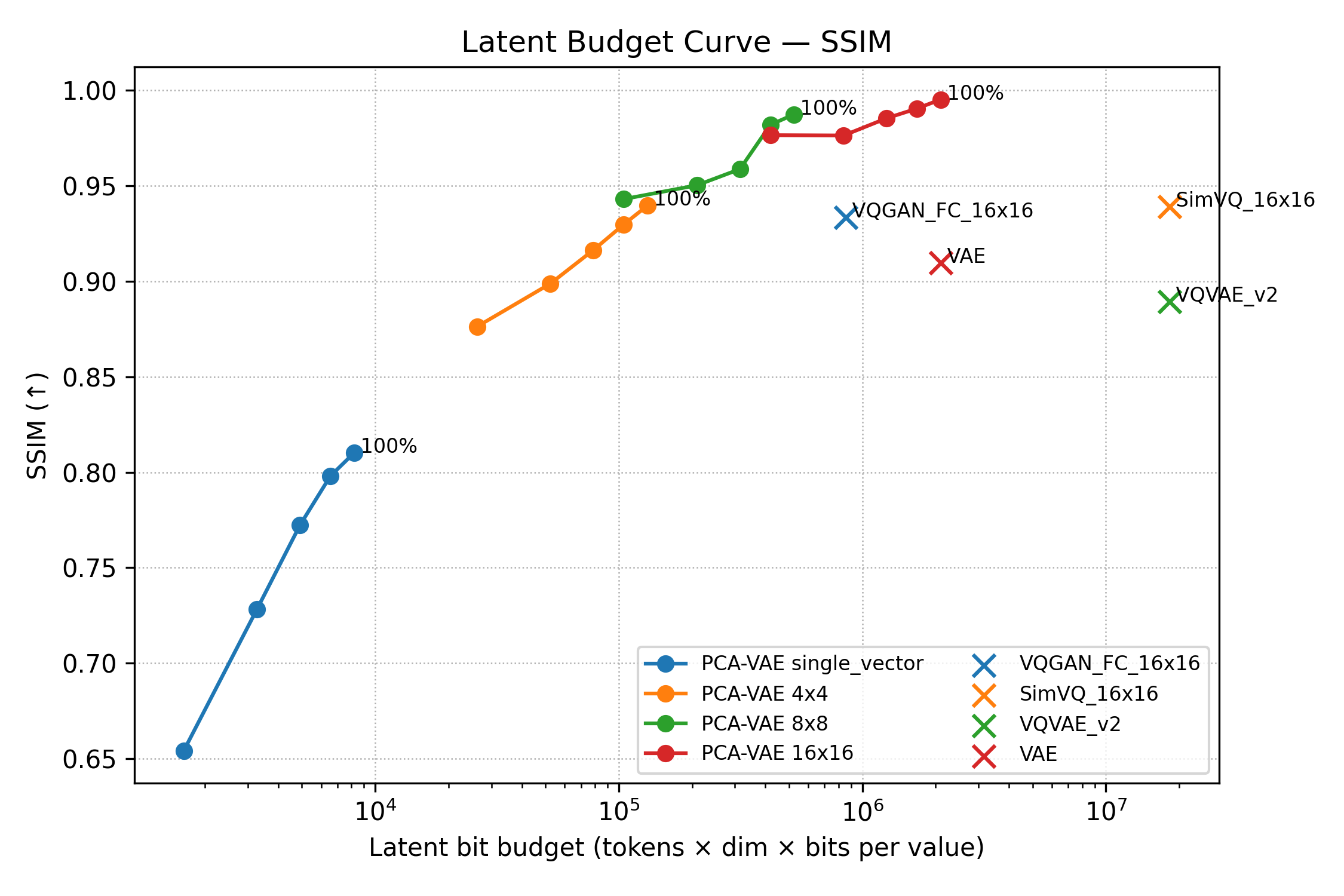}
    \caption{SSIM (higher is better)}
\end{subfigure}
\caption{
Latent bit–budget curves comparing PCA–VAE to VQGAN~\cite{RN3}, SimVQ~\cite{RN13}, VQ-VAE~\cite{razavi2019generating}, and AutoencoderKL~\cite{RN4}. 
PCA–VAE achieves higher reconstruction quality per bit across PSNR, SSIM, LPIPS, and rFID, 
often matching VQ performance with 10$\times$-100$\times$ fewer latent bits. 
}
\label{fig:latent_budget}
\end{figure*}

Figure~\ref{fig:latent_budget} reports PSNR, SSIM, LPIPS, and rFID as a function
of the latent bit budget across PCA-VAE configurations (1$\times$1, 4$\times$4,
8$\times$8, and 16$\times$16 latent grids) and VQ baselines. PCA-VAE exhibits
consistent quality improvements as additional principal components are included,
forming a smooth scaling curve. Remarkably, PCA-VAE achieves comparable or better
reconstruction quality than VQGAN~\cite{RN3}, SimVQ~\cite{RN13}, and VQ-VAE~\cite{razavi2019generating} while using
one to two orders of magnitude fewer latent bits. For example, the 8$\times$8
PCA-VAE configuration reaches SimVQ~\cite{RN13}--level PSNR and SSIM despite consuming
approximately $10\times$--$30\times$ fewer bits, and the 16$\times$16
PCA-VAE variant attains near--zero LPIPS and rFID at significantly lower budgets
than any discrete tokenizer.

These results demonstrate that continuous orthogonal latent representations
achieve higher information density than discrete codebooks.
Whereas VQ models require large token streams to capture perceptual structure,
PCA-VAE concentrates signal energy into a compact set of principal axes,
yielding higher reconstruction quality per bit and establishing a favorable
bit--quality scaling law for orthogonal latent spaces.

\subsection{Latent Factor Interpretability}

\begin{figure}[t]
    \centering
    \includegraphics[width=\linewidth]{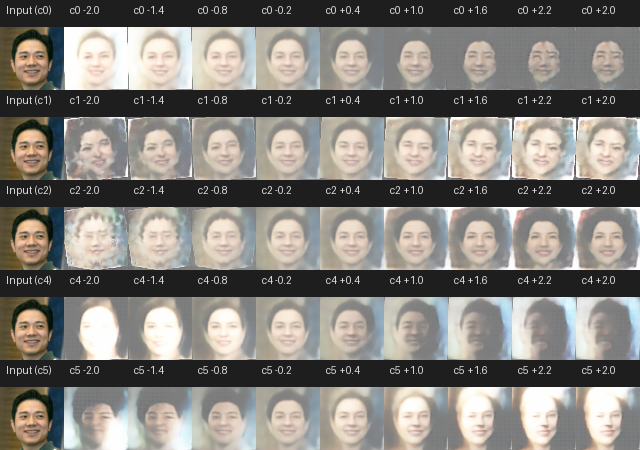}
    \caption{
    Latent semantics in PCA-VAE.
    We modify a single latent coefficient within the range $[-2, 2]$
    while fixing the others. Each principal direction produces a
    coherent semantic transition (illumination, head pose, facial structure,
    shading, hair density), demonstrating interpretable continuous latent axes.
    }
    \label{fig:latent_semantics}
\end{figure}

Orthogonal latent representations naturally induce disentangled semantic axes.
To examine the interpretability of PCA-VAE latents, we perform controlled latent
perturbation experiments on a held--out face image. We use the 1$\times$1
PCA-VAE, identify its first five principal components, and independently vary
each latent coefficient within the range $[-2, 2]$ while holding all others
fixed. Figure~\ref{fig:latent_semantics} visualizes the resulting reconstructions.

Each principal direction produces a coherent and consistent semantic change.
The first component adjusts global illumination (dark--to--bright background),
the second induces head pose rotation, the third smoothly transitions facial
structure from masculine to feminine, the fourth modulates facial shading, and
the fifth alters hair density (from thick hair to near baldness). These
effects remain stable across the perturbation range and do not collapse into
texture noise or incoherent artifacts.

Unlike discrete codebook latents, which lack a natural coordinate geometry,
PCA-VAE yields continuous, ordered, and interpretable latent axes due to its
orthogonal construction. This structure enables semantic manipulation without
adversarial latent directions or iterative search, demonstrating that PCA-VAE
possesses inherent latent organization typically attributed to specialized
disentanglement objectives.

Unlike VAE latents, which lack a canonical ordering, PCA-VAE automatically
sorts latent dimensions by information content, concentrating salient semantic
factors in the first few coordinates. This yields interpretable and
hierarchically organized latent axes, enabling both compact representation and
direct semantic manipulation.

Across all experiments, PCA-VAE demonstrates three key advantages. 
First, it consistently achieves higher reconstruction quality per latent bit, 
revealing a favorable scaling law and superior information efficiency compared to 
discrete tokenizers. Second, its latent dimensions form ordered, orthogonal, 
and interpretable semantic axes, enabling direct and compact control of perceptual 
attributes. Third, PCA-VAE exhibits smooth, monotonic performance gains as the 
number of retained principal components increases, offering predictable 
capacity–quality trade-offs without codebook collapse or instability. 
Together, these findings indicate that orthogonal continuous latent spaces 
provide a simple yet powerful alternative to learned vector quantization.

\section{Discussion}

Vector-quantized autoencoders require non-differentiable codebook lookup and rely on 
straight-through estimators to enable training. They may also suffer from codebook 
collapse when the codebook size increases. 
PCA-VAE naturally avoids both issues: its latent transformation is fully differentiable, 
no surrogate gradients are required, and no codebook exists to collapse. Despite this 
simplicity, PCA-VAE achieves strong reconstruction performance and favorable bit–quality 
scaling behavior, rivaling or exceeding state-of-the-art VQ methods.

PCA-VAE inherits the orthogonality and spectral ordering of principal components, 
producing a latent space in which axes are automatically ranked by explained variance. 
This yields semantically meaningful directions concentrated in the leading components, 
unlike conventional VAE latents, which lack a canonical ordering or guaranteed disentanglement. 
Such a structure provides both interpretability and efficient compression, and it suggests 
a promising inductive bias for representation learning.

The PCA layer used in PCA-VAE is a linear transformation and therefore modular: 
it can be inserted into existing neural architectures to induce interpretability 
and rank-aware latent organization. This suggests opportunities beyond 
autoencoding, including generative modeling, vision transformers, 
and multimodal encoders, where latent structure and controllability are desirable.

\section{Conclusion}
We introduced PCA-VAE, a variational autoencoder that replaces discrete codebook 
quantization with an orthogonal principal component layer. This design yields 
interpretable and hierarchically ordered latent axes while achieving superior 
information efficiency and strong reconstruction quality. PCA-VAE provides 
a simple, differentiable alternative to vector quantization, eliminating the need 
for straight-through estimators and avoiding codebook collapse.

Our study focuses on reconstruction quality, and we evaluate primarily on 
CelebA--HQ at 256$\times$256 resolution. We have not yet explored generative 
sampling or scaling to large datasets and models.

Future directions include extending PCA-VAE to fully generative models, 
training on large-scale datasets, and investigating PCA-based latent layers 
as a general building block to improve interpretability and controllability 
across diverse neural architectures.

{
    \small
    \bibliographystyle{IEEEtran}
    \bibliography{main}
}
\clearpage
\setcounter{page}{1}
\maketitlesupplementary
\section*{Additional Qualitative Results}

\begin{figure*}[h]
\centering
\includegraphics[width=0.7\linewidth]{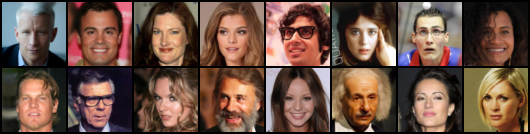} \\
\vspace{2pt}
\small Ground Truth
\vspace{6pt}

\includegraphics[width=0.7\linewidth]{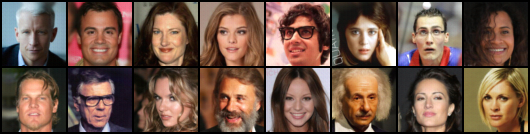} \\
\vspace{2pt}
\small PCA-VAE (ours)

\vspace{6pt}
\includegraphics[width=0.7\linewidth]{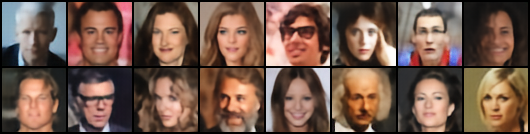} \\
\vspace{2pt}
\small AutoencoderKL (Stable Diffusion VAE baseline)

\vspace{6pt}
\includegraphics[width=0.7\linewidth]{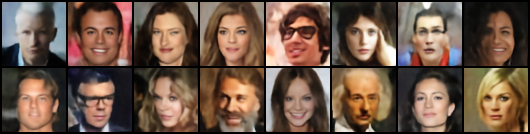} \\
\vspace{2pt}
\small VQ-VAE v2

\vspace{6pt}
\includegraphics[width=0.7\linewidth]{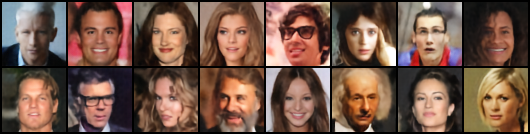} \\
\vspace{2pt}
\small SimVQ (strong VQ baseline)

\vspace{6pt}
\includegraphics[width=0.7\linewidth]{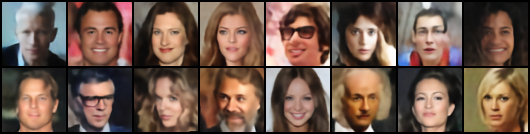} \\
\vspace{2pt}
\small VQGAN-FC

\caption{
Qualitative reconstruction comparison on CelebA-HQ (256$\times$256). 
Each row displays reconstructions from a different model using the same set of input images (top row). 
PCA-VAE produces sharp and identity-preserving reconstructions, noticeably improving facial structure (e.g., eye and mouth shapes) and global consistency compared to AutoencoderKL and VQ-VAE v2. 
SimVQ and VQGAN-FC perform competitively but occasionally exhibit blocky textures or over-smoothed regions, whereas PCA-VAE achieves similarly strong fidelity without vector quantization, codebook lookup, or STE.
}
\label{fig:supp_recon}
\end{figure*}

To further validate the reconstruction advantages of PCA-VAE, 
we provide qualitative visualizations on held-out CelebA-HQ images 
(Fig.~\ref{fig:supp_recon}). 
The compared methods include:

\begin{itemize}
\item Ground-truth reference images
\item PCA-VAE (ours)
\item AutoencoderKL (Stable Diffusion VAE)
\item VQ-VAE v2
\item SimVQ (strong VQ baseline)
\item VQGAN-FC
\end{itemize}

All models are trained on CelebA-HQ at 256$\times$256 resolution.  
The same inputs are shown for every method. 
No test-set overlap or cherry-picking was performed.
 
Across identities, poses, and lighting conditions, PCA-VAE produces 
clearer and more faithful reconstructions relative to VAE and VQ baselines.  
We highlight the following consistent effects:

\begin{itemize}
\item Sharper facial features, especially around eyes, nose, and mouth
\item Cleaner textures in skin, hair, and background regions
\item Better identity preservation, including face shape and hairstyle
\item Fewer artifacts than VQ models (no patch blocking or codebook noise)
\item No over-smoothing, a common failure mode in AutoencoderKL
\end{itemize}

Visually, PCA-VAE resembles the fidelity of the best VQ models (SimVQ / VQGAN-FC) 
while avoiding their quantization artifacts and without requiring a codebook.

This supports our claim that PCA-VAE provides:
(1) competitive or superior perceptual quality, and  
(2) a more stable and interpretable latent space.

\section*{Latent Semantics and Emergent Correlations}

\begin{figure}[h]
\centering
\includegraphics[width=\linewidth]{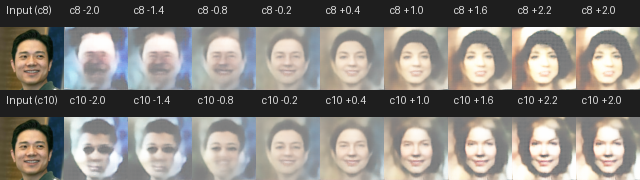}
\caption{
Latent interpolation along the 8th and 10th PCA dimensions. 
Although many dimensions capture well-defined semantics, 
some attributes co-vary (e.g., emotion and gender appearance, 
sunglasses and perceived masculinity). 
This reveals interpretable yet not strictly disentangled latent factors, 
likely reflecting data co-occurrence structure rather than model artifacts.
}
\label{fig:supp_latent_bias}
\end{figure}

Beyond controllable attribute directions shown in the main paper, 
we further explore additional PCA-VAE latent dimensions by applying 
scalar perturbations to individual components. 
Fig.~\ref{fig:supp_latent_bias} visualizes interpolations on the 
8th and 10th principal dimensions for the same input identity.

Interestingly, we observe that semantic factors are not always 
perfectly disentangled across dimensions. 
Certain attributes emerge in correlated groups. 
For example:

\begin{itemize}
\item Negative shifts on $c_{8}$ tend to produce a sad expression 
  together with a more masculine facial structure.
\item Positive shifts on $c_{8}$ often produce a smiling expression 
  together with a feminine facial structure.
\item Perturbing $c_{10}$ yields similar co-occurrence patterns: 
  sunglasses and a confident pose often appear alongside more masculine features,
  while removing glasses and softening expression often correlate with 
  more feminine facial traits.
\end{itemize}

These phenomena suggest that while PCA-VAE latents are ordered by 
variance and exhibit strong semantic interpretability, 
complete disentanglement is not guaranteed. 
Instead, certain emerging factors reflect natural co-occurrence patterns 
in the training distribution, revealing intrinsic dataset and 
representation biases.

\end{document}